# A Portfolio Approach to Algorithm Selection for Discrete Time-Cost Trade-off Problem

Santosh Mungle

*Department of Systems and Industrial Engineering, University of Arizona, USA - 85721-0020*

## Abstract

It is a known fact that the performance of optimization algorithms for NP-Hard problems vary from instance to instance. We observed the same trend when we comprehensively studied multi-objective evolutionary algorithms (MOEAs) on a six benchmark instances of discrete time-cost trade-off problem (DTCTP) in a construction project. In this paper, instead of using a single algorithm to solve DTCTP, we use a portfolio approach that takes multiple algorithms as its constituent. We proposed portfolio comprising of four MOEAs, Non-dominated Sorting Genetic Algorithm II (NSGA-II), the strength Pareto Evolutionary Algorithm II (SPEA-II), Pareto archive evolutionary strategy (PAES) and Niched Pareto Genetic Algorithm II (NPGA-II) to solve DTCTP. The result shows that the portfolio approach is computationally fast and qualitatively superior to its constituent algorithms for all benchmark instances. Moreover, portfolio approach provides an insight in selecting the best algorithm for all benchmark instances of DTCTP.

## 1. Introduction

The discrete time-cost trade-off problem (DTCTP) is an important aspect in the scheduling of construction project. In project planning and scheduling, selection of appropriate resources, including crew sizes, equipment, methods and technologies to carry out a project are challenging decisions. These decisions ultimately affect the duration and cost of the project. In practice, if the project is running behind the scheduled plan, acceleration of activity becomes essential by allowing additional resources which ultimately incurs an extra cost [1]. DTCTP is a process to identify a suitable activity for speeding up and deciding "by how much" so as to attain the best possible savings in both time and cost [2]. Such optimization process includes hidden trade-off relationship between time and cost. This hidden trade-off between time and cost creates a complicated situation for project manager's to predict whether the total cost (*i.e.*, the direct and indirect costs) would increase or decrease as a result of the schedule compression.

Over the last few decades, various approaches have been proposed to solve DTCTP. They can be broadly classified based on mathematical models and metaheuristics. Researchers have developed various mathematical models using linear programming, integer programming, and dynamic programming to solve a small size of problem instances of DTCTP [3-5]. However, mathematical models require great computational effort, and some approaches do not provide the optimal solution either for large size instances of DTCTP. Moreover, DTCTP is known as the NP-hard problem that failed to maintain pseudo-polynomial time guarantee, for complex and large project network using exact solution algorithms [1]. Therefore, recently, many researchers proposed various metaheuristics such as genetic algorithm (GA), ant colony optimization (ACO), simulated annealing (SA) etc. to solve the DTCTP for large-scale project networks to obtain near-optimal Pareto solution within a reasonable timeframe [6-13]. However, it is not easy to inform in advance for decision makers, which metaheuristic is the best to solve an instance of NP-hard optimization problem. Since a behavior of most metaheuristic algorithms for NP-hard problems are usually tough to characterize analytically and experimentally that further lead to the problem of algorithm selection [14]. Previously, researchers have made few attempts to solve algorithm selection problem for NP-hard problem by using experimental methods [15, 16]. Moreover, for large-scale networks (several hundred activities with discrete time-cost relationships, which are common for most construction projects), neither heuristic methods nor mathematical models can offer optimal solutions efficiently [11].

DTCTP is a multi-objective optimization problem that attempts to strike a delicate balance between project schedule time and cost. A multi-objective optimization problem, in its real sense, requires the determination of a representative set of non-dominated or Pareto optimal solutions in the stipulated time frame. The last decade has witnessed a significant growth in the use of MOEAs for complex multi-objective optimization problems [17]. Continuous improvements in the past few years have spectacularly reduced the time of response of these meta-heuristics without much depreciation regarding solution quality. In particular, MOEAs such as Non-dominated Sorting Genetic Algorithm II (NSGA-II) [18], Strength Pareto Evolutionary Algorithm II (SPEA-II) [19] and Niched Pareto Genetic Algorithm II (NPGA-II) [20] have been extensively used to solve complex multiobjective optimization problems. It's mainly because of their robustness and capability to handle a large-scale problem size of multiobjective optimization.

The objective of this paper is to design a portfolio of four MOEAs namely NSGA-II [18], SPEA-II [19], NPGA-II [20] and PAES [21], and its implementation on DTCTP. These four MOEAs are parallelly processed on two and four processors system. Moreover, the average quality metric is used to evaluate the quality of the approximated Pareto-optimal front. The average quality metric aids the decision-making process to investigate the relative performance of various strategies embedded in the portfolio to get an insight into the solution quality of the Pareto-optimal front with the increasing problem size and complexity. A portfolio of algorithms is formally expressed as "*A collection of different algorithms and or different copies of the same algorithm running on different processors*" [22-24]. Algorithm portfolio seeks to identify the best-suited strategy and thereby amalgamate the performance of various algorithms, thus enhancing the algorithm performance in a dynamic environment. This effect can be termed as a performance-maximizing superposition of algorithms where an algorithm performance is tailored for the performance of counterparts embedded in the portfolio. Finally, AHP method is adapted to determine the best portfolio case for DTCTP.

The remainder of the paper is organized as follows: Section 2 presents the mathematical formulation of DTCTP. Section 3 reviews some of the terminologies and techniques employed in the field of multi-objective optimization. Section 4 presents the computational study. Section 5 presents the design of algorithm portfolio and its implementation. Section 6 includes the numerical results. Finally, section 7 concludes the paper with suggested future research works.

**2. Mathematical Formulation**

To model the DTCTP, we start with the standard assumption for modelling the project network, as follows: 1) The project network has no cycles; 2) The start activity (activity 0, a dummy activity) is the only activity that is not an immediate successor of any activity; 3) The finishing activity (activity $N+1$, also a dummy activity) is the only activity that has no successors.

In DTCTP, project completion time, and implementation cost are primary objectives in determining the best possible option for every activity in the project network.

## 2.1. Project completion time

The project completion time is the sum of the durations of scheduled activities in the project. It can be expressed, as follows:

$$\min \left\{ \max_{L_k \in L} \sum_{i \in L_k} \sum_{j=1}^{n_i} t_{ij} x_{ij} \right\} \tag{1}$$

Where,

$t_{ij}$ = Duration of activity $i$ for option $j$, for $i = 1, \ldots, N, j = 1, \ldots, n_i$

$N$ = Total number of activities in the project network

$x_{ij}$ = If $x_{ij} = 1$ then activity $i$ perform $j^{th}$ option, while $x_{ij} = 0$ means not

$L_k$ = Activity sequence of $k^{th}$ path $(ij_k = i1_k, i2_k, \ldots, in_k)$

$L$ = Sets of all path of a network $(1, 2, \ldots, m)$

$m$ = Number of paths in a project network

$n_i$ = Number of time − cost option for each activity $i$, for $i = 1, \ldots, N$

## 2.2. Project cost

The total project cost consists of direct and indirect cost. The direct cost is the sum of the direct cost of all activities within a project network. On the other hand, indirect cost comprises of the management expenditure during project implementation, which depends heavily on the project duration, i.e., longer the duration, the higher the indirect cost. The total project cost in mathematical form is, as follows:

$$\min \sum_{i=1}^{N} \sum_{j=1}^{n_i} C_{ij} x_{ij} + \min \left\{ \max_{L_k \in L} \sum_{i \in L_k} \sum_{j=1}^{n_i} t_{ij} x_{ij} \right\} \times IC_{ij} \tag{2}$$

Where,

$C_{ij} = m_{ij} + t_{ij} R_{ij}$

$R_{ij}$ = Daily cost rate in $\frac{\$}{day}$ for using option $j$ in activity $i$, for $i = 1, \ldots, N, j = 1, \ldots, n_i$

$C_{ij}$ = Direct cost of option $j$ for activity $i$, for $i = 1, \ldots, N, j = 1, \ldots, n_i$

$IC_{ij}$ = Indirect cost of option $j$ for activity $i$, for $i = 1, \ldots, N, j = 1, \ldots, n_i$

$m_{ij}$ = Material cost of option $j$ for activity $i$, for $i = 1, \ldots, N, j = 1, \ldots, n_i$

$t_{ij}$ = Duration of activity $i$ for option $j$, for $i = 1, \ldots, N, j = 1, \ldots, n_i$

### 2.3. Constraints

#### 2.3.1. Network logic constraints

The project network is always constrained by one of two methods [25]. The first method allows precedence constraints for each immediate preceding relationship in the project network. The second method allows one constraint for each path from the first activity to the last one in the project network. In our mathematical model, the first method is adopted. The logical relationship between any two consecutive activities $i$ and its immediate successor $k$ is expressed as:

$$SS_k \geq SF_i \qquad \forall i = 1, \ldots, N, \ \forall k \in S_i \tag{3}$$

Where,

$SS_k$ = Scheduled start of immediate succesor of activity $i$

$SF_i$ = Scheduled finishing time of activity $i$

The $SF_i$ equal scheduled start time ($SS_i$) plus its duration. The logical relationship constraint can be expressed using following equation, in which $S_i$ is the set of successor activities to activity $i$.

$$SF_k \geq SF_i + \sum_{j=1}^{n_i} t_{ij} x_{ij} \qquad \forall i = 1, \ldots, N, \forall k \in S_i \tag{4}$$

#### 2.3.2. Project completion constraints

In real life scenario, we apply a constraint on project completion time and cost to ensure that the both the objective will complete under restriction. These constraints are expressed by the following equations, in which $T_{max}$ is denoted as a maximum allowable project duration and $C_{max}$ is denoted as a maximum allowable project cost.

$$\min \left\{ \max_{L_k \in L} \sum_{i \in L_k} \sum_{j=1}^{n_i} t_{ij} x_{ij} \right\} < T_{max} \tag{5}$$

$$\min \sum_{i=1}^{N} \sum_{j=1}^{n_i} C_{ij} x_{ij} + \min \left\{ \max_{L_k \in L} \sum_{i \in L_k} \sum_{j=1}^{n_i} t_{ij} x_{ij} \right\} \times IC_{ij} < C_{max} \tag{6}$$

## 3. Basic Concepts of Multi-objective Optimization

In this section, we briefly present the basic concepts of multi-objective optimization. Multi-objective optimization problem can be formulated as follows:

$$\text{maximize } \{f_1(x) = z_1, \ldots, f_J(x) = z_J\} \tag{7}$$
$$s.t. \qquad x \in X$$

Where, solution $x = [x_1, \ldots, x_I]$ is a vector of decision variables in a problem at hand and $X$ is the set of feasible solutions available in a search space. If the variables in a problem at hand are discrete, the multi-objective optimization problem is called multiple-objective combinatorial optimization problems.

The image of a solution $x$ in the objective space is a point

$$z^x = [z_1^x, \ldots, z_J^x] = f(x), \text{ such that } z_j^x = f_j(x) \quad j = 1, \ldots, J \tag{8}$$

Point $z^1$ dominates $z^2$, $z^1 \succ z^2$, if $\forall j \; z_j^1 \geq z_j^2$ and $z_j^1 > z_j^2$ for at least one $j$. Solution $x^1$ dominates $x^2$ if the image of $x^1$ dominates the image of $x^2$. A solution $x \in X$ is considered as Pareto-optimal if there is no $x' \in X$ that dominates solution $x$. A point being an image of a Pareto-optimal solution is known as a non-dominated point. The set of all Pareto-optimal solutions is called a Pareto-optimal set. The image of the Pareto-optimal set in the objective space is called the non-dominated set or Pareto front. An approximation of the non-dominated set is a set $A$ of feasible points (and corresponding solutions) such that $z^1, z^2 \in A$ such that $z^1 \succ z^2$.

The point $z^*$ composed of the best attainable objective function value is called the ideal point.

$$z_j^* = max\{f_j(x) | x \in X\}, \qquad j = 1, \ldots, J \tag{9}$$

An approximation of the ideal point based on set $A$ is denoted by $\mathbf{Z}^{**}(A)$

$$z_j^{**}(A) = \max\{z_j | z \in A\}, \qquad j = 1, \ldots, J \tag{10}$$

Finding the full Pareto front is a computationally tough task which is attributed to the presence of a vast number of sub-optimal Pareto front. Considering a memory limitation, determining the full Pareto front becomes infeasible, and thus requires the Pareto front to be diverse covering maximum possible regions of it. Therefore, a multi-objective algorithm aims to obtain an approximation of the Parent front [26].

In this paper, the performance of algorithm portfolio is measured based on the quality of the obtained Pareto front, which is quantitatively evaluated using standard criteria or metrics. We utilized one such metric 'Average Quality' (AQ) [26] here. In the past, the quality of the Pareto-optimal set was measured using Tchebycheff function $S_\infty$ [26], as follows:

$$S_\infty(\mathbf{z}, \mathbf{z}^0, \wedge) = \max_j \left\{ \lambda_j (z_j^0 - z_j) \right\} \\ = \max_j \left\{ \lambda_j \left( z_j^0 - f_j(\mathbf{x}) \right) \right\} \tag{11}$$

Where $z^0$ is a reference point, $\wedge = [\lambda_1, \ldots, \lambda_j]$ is a weight vector such that $\lambda_j \geq 0 \ \forall j$ and $\exists j | \lambda_j > 0$. Each weighted Tchebycheff scalarizing function has at least one global optimum belonging to the set of Pareto-optimal solutions. Note, however, that if the optimum is not unique, then some of the optima may be dominated, but must have at least one objective component equal to a Pareto-optimal solution. For each Pareto optimal solution **x** there exists a weighted Tchebycheff scalarizing function $S$ such that *x* is a global optimum of *S*.

However, this function tends to hide certain aspects regarding the solution set's quality because poor performance regarding proximity may get compensated by good performance in the distribution of solutions. To overcome this limitation, diversity indicators of spread and space are added to the formulation of AQ metric.

Achievement scalarizing function is defined in the following ways:

$$S_a(\mathbf{z}, \mathbf{z}^0, \wedge, \rho) = \max_j \left\{ \left\{ \lambda_j (z_j^0 - z_j) \right\} + \rho \sum_{j=1}^{J} \lambda_j (z_j^0 - z_j) \right\} \tag{12}$$

Where $\rho$ = small positive number, $0 < \rho << 1$. All global optima of each achievement scalarizing function belong to the set of Pareto-optimal solutions.

Weight vector that meets the following condition is called normalized weight vectors.

$$\forall j \lambda_j \geq 0, \quad \sum_{j=1}^{J} \lambda_j = 1 \tag{13}$$

The AQ metric uses a representative sample $\Psi_s$ of normalized weight vectors which is represented as:

$$\Psi_S = \left\{ \Lambda = [\lambda_1, \ldots, \lambda_J] \in \Psi \mid \lambda_j \in \left\{ 0, \frac{1}{l}, \frac{2}{l}, \ldots, \frac{l-1}{l}, 1 \right\} \right\} \tag{14}$$

Where,

$\Psi_s$ = Set of normalized vector

$l$ = Sampling parameter

We observe that $\Psi_s$ contain $\binom{l+J-1}{J-1}$ weight vectors. For each number of objectives $J$, the size of $\Psi_s$ should at least 50 *i.e.,* sampling parameter $l$ is set as the lowest value that assures $|\Psi| \geq 50$. Therefore $l$ set to be 49 for $J = 2$. Each weight vector $\Lambda \in \Psi_s$ defines an achievement scalarizing function $S_a(z, z^0, \Lambda, \rho)$. All scalarizing functions defined by vectors from set $\Psi_s$ constitute set $S_a$ of functions.

To evaluate the quality of solution generated by algorithm portfolio, we first run the algorithm portfolio. In result, an approximation $A$ of the whole non-dominated set is obtained. Then for each function $S_a(z, z^0, \Lambda, \rho) \in S_a$ the best solution $x$ on this function is selected from set $A$. i.e., $\forall z \in A \ S_a(f^x, z^0, \Lambda, \rho) \leq S_a(z, z^0, \Lambda, \rho)$. Thus the average quality of solutions generated by the portfolio is

$$AQ = \frac{\sum_{A \in \Psi_S} S_a(f^x, z^0, \Lambda, \rho)}{|\Psi_S|} \tag{15}$$

Where,

$Z^0$ = reference point which is an approximation of the ideal point
$X$ = best solution on function $S_a(z, z^0, \Lambda, \rho) \mid x \in A$
$\rho$ = parameter set as 0.01

For each test instance, an approximation of the ideal point was found by local optimization of individual objectives started from an initial solution. This approximation of the ideal point is used as reference point in achievement scalarizing function $S_a(z, z^0, \Lambda, \rho)$. The objective ranges observed during local optimization of every individual objective is used to determine the range equalization factor which further helps to normalize the objective values before calculating scalarizing function.

The range equalization factors [27] are defined in the following ways:

$$\pi_j = \frac{1}{R_j}, \qquad j = 1, \dots, J \qquad (16)$$

Where, $R_j$ is the approximate range of objective function $Z_j$ in the non-dominated set $A$. Objective values multiplied by the range equalization factors are called normalized objective function values.

## 4. Computational Study

### 4.1. Benchmark instances of DTCTP

In this paper, benchmark instances of DTCTP are characterized based on the number of activities, possible schedules, and a number of total paths in the project network. Table 1 shows the benchmark instances of DTCTP with their characteristics above. In our computational study, total six benchmark instances of DTCTP of the size of very large to very small have been considered. Among them, problem instances 1, 2, 3, 5 and 6 have been adapted from [11], [7], [8], [12] and [13], and problem 4 is hypothetically generated with maximum 4 and minimum 2 execution modes for project activities. The solution to these problem instances has been attempted by integrating them into an algorithm portfolio that works on a strategy of minimizing the risk regarding computational cost and the solution quality obtained.

**Table 1**: Benchmark Instances

| Sr. No. | Number of Activities | Possible Schedules | Number of Network Paths |
|---|---|---|---|
| 1 | 63 | $1373*10^{42}$ | 28 |
| 2 | 29 | $8264*10^{6}$ | 46 |
| 3 | 18 | $1494*10^{7}$ | 11 |
| 4 | 14 | $2831*10^{4}$ | 11 |
| 5 | 9 | $1500*10^{3}$ | 5 |
| 6 | 7 | 5569 | 3 |

## 4.2. Multi-objective Evolutionary Algorithms (MOEAs)

In this paper, four establish MOEAs namely, NSGA-II [18], SPEA-II [19], NPGA-II [20] and PAES [21] is taken into consideration to design and implement the algorithm portfolio. The following discussion presents a brief overview of these four MOEAs.

### 4.2.1. Non-dominated Sorting Genetic Algorithm-II (NSGA-II)

It was proposed by Deb et al. [18]. This algorithm is a revised version of Non-dominated Sorting Genetic Algorithm (NSGA) proposed by Srinivas and Deb [28] which has been criticized mainly for its high computational complexity, non-elitism approach and a need for specifying a sharing parameter. NSGA was based on a Genetic Algorithm (GA), which utilizes several layers of classification of the individuals. NSGA-II follows a fast non-dominated sorting approach which requires O ($MN^2$) comparisons. NSGA-II also replaced the use of sharing function with the newly crowded comparison approach that eliminates the need for any user-defined parameter for maintaining diversity among population members. The other basic steps of NSGA-II are same as those of GA. Initially, a population double in size ($2N$) is randomly generated and is then sorted by non-dominance. After that, the selection of best $N$ members from the population is carried out. Crossover and Mutation are performed to utilize these genetic operations to produce better offspring. The parents and offspring are then combined to form the initial population for the next generation and the aforementioned procedure is repeated. The algorithm finally provides a well distributed Pareto set of solutions. In the present paper, two point crossover and bit-flip mutation have been used.

### 4.2.2. Strength Pareto Evolutionary Algorithm II (SPEA-II)

It is a revised version of SPEA proposed by Zitzler and Thiele [19]. SPEA was developed using a previously carried comparative study. It establishes the multi-objective techniques commonly employed in Evolutionary Algorithms (EAs) into a single meta-heuristic skeleton [29]. SPEA-II utilizes an external memory for storing the non-dominated individuals. Each individual in the extended memory as well as in the current population is assigned a strength value based on the domination and density information. The rank value of an individual is determined by the summation of strength values of the individuals that dominate it. The density of each individual is estimated based on the $k^{th}$ nearest neighbor density estimation method. The final fitness value is then calculated as the sum of the rank and density values. In addition, a hierarchical clustering based method is adopted for the pruning of the external archive thereby maintaining diversity in the obtained Pareto front [30].

### 4.2.3. Niched Pareto Genetic Algorithm-II (NPGA-II)

Erickson et al. [20] proposed a revised version of the NPGA [31] called the NPGA-II. This algorithm uses Pareto ranking, but keeps the tournament selection (solving ties through fitness sharing as in the original NPGA). In this case, no external memory is used and the elitist mechanism is similar to the one adopted by the NSGA-II. Niche counts in the NPGA-II are calculated using individuals in the partially filled next generation, rather than using the current generation. This is called continuously updated fitness sharing and was proposed by Oei et al. [32].

### 4.2.4. Pareto Archived Evolution Strategy (PAES)

This algorithm was introduced by Knowles and Corne [21]. PAES consists of a (1 + 1) evolution strategy (*i.e.*, a single parent that generates a single offspring) in combination with a historical archive that records the non-dominated solutions previously found. This archive is used as a reference set against which each mutated individual is being compared. Such a historical archive is the elitist mechanism adopted in PAES. However, an interesting aspect of this algorithm is the procedure used to maintain diversity which consists of a crowding procedure that divides objective space in a recursive manner. Each solution is placed in a certain grid location based on the values of its objectives (which are used as its "coordinates'' or "geographical location''). A map of such grid is maintained, indicating the number of solutions that reside at each grid location. Since the procedure is adaptive, no extra parameters are required (except for the number of divisions of the objective space).

### 4.3. Computational Analysis

As we know that MOEAs do not always reach the optimal solution, even for long computation times. In addition, it is often impossible to obtain an analytical prediction of either the solution achievable within a given computation time or the time is taken to find a solution of a given quality. The analysis of these measures is critical to the comparison between MOEAs. In literature, statistical analysis of these measures is known as "Univariate Model" [33]. More specifically, Univariate model is the one in which either the solution cost or run-time is taken into consideration while analyzing the various metaheuristics.

In this paper, our concern is run-time that means computation time is measured when a solution with the desired property is found. We assume that run-time corresponds to the number of iterations needed to achieve the desired solution quality.

The performance measure $X$ (run-time) of metaheuristics on the instances can be described by a probability distribution function $f(x) = P[X = x]$ or equivalently by its cumulative (discrete) distribution function (CDF) $f(x) = P[X \leq x] = \sum_{x_i \leq x} p(x_i)$

In the computational experiments, we observe data $x_1, x_2, \ldots, x_n$ and then calculate the CDF. Once the CDF is known, parameter such as the mean and variance of the iterations is calculated to compare the performance of each MOEA.

Initiation of the experiment takes place by testing the four MOEAs on all the benchmark instances for 50 trials and the results have been reported in Figures 1-6. In Figure 1-6, $X$-axis represents the iterations needed to satisfy the termination criteria and $Y$-axis represents the CDF value. The termination criteria for each MOEA set at the performance level of 10% of the best-obtained AQ (average quality) metric value. The mean and variance of the number of iterations required by each algorithm for the six problems is provided in Table 2.

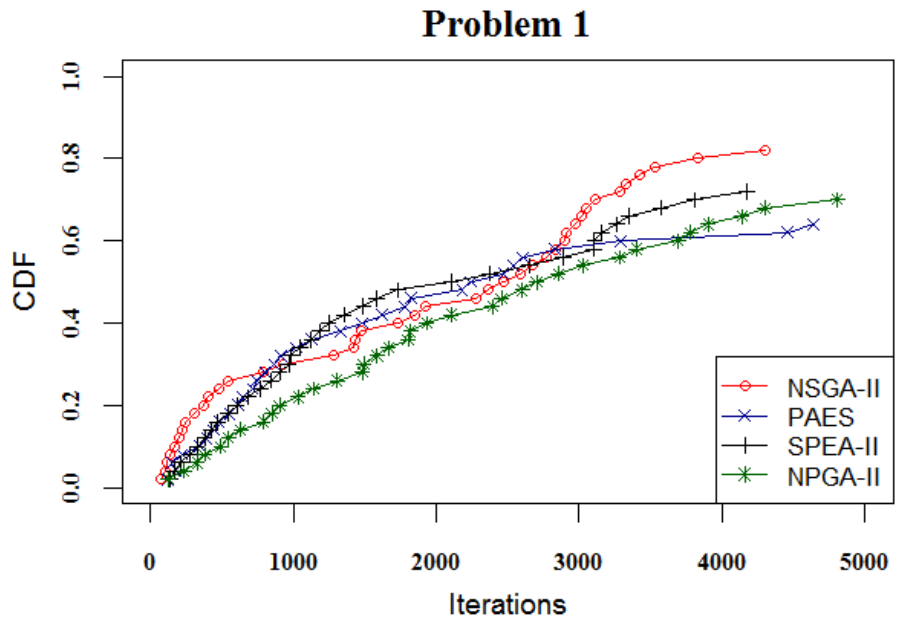

**Figure 1:** CDF of 4 MOEAs for Problem 1

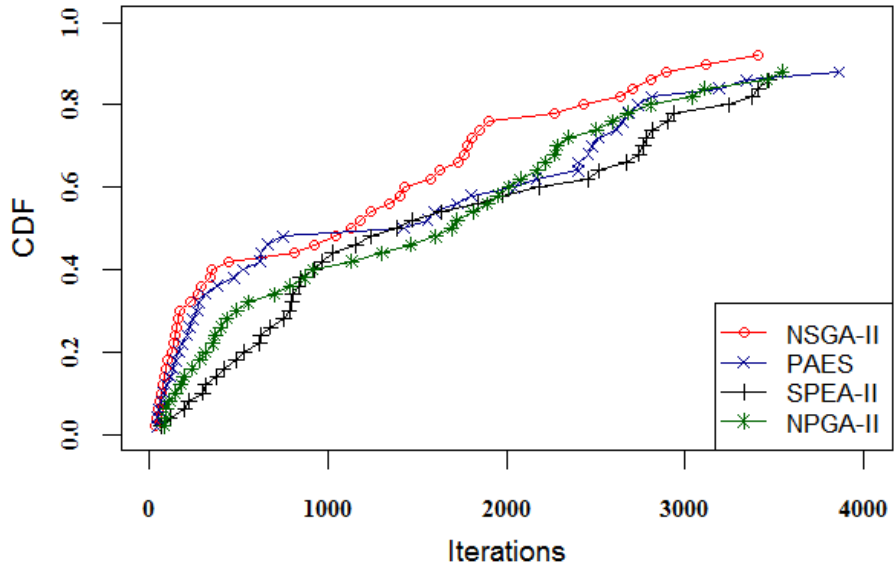

**Figure 2:** CDF of 4 MOEAs for Problem 2

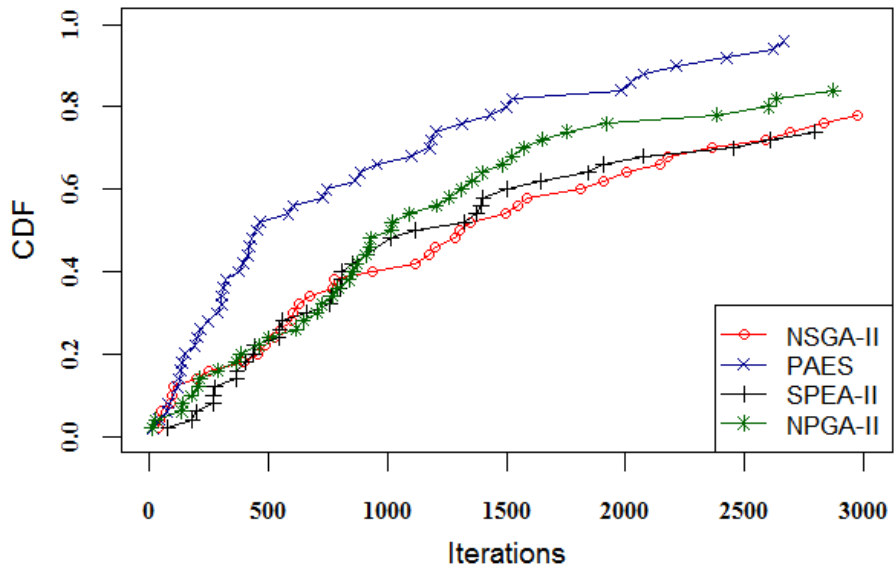

**Figure 3:** CDF of 4 MOEAs for Problem 3

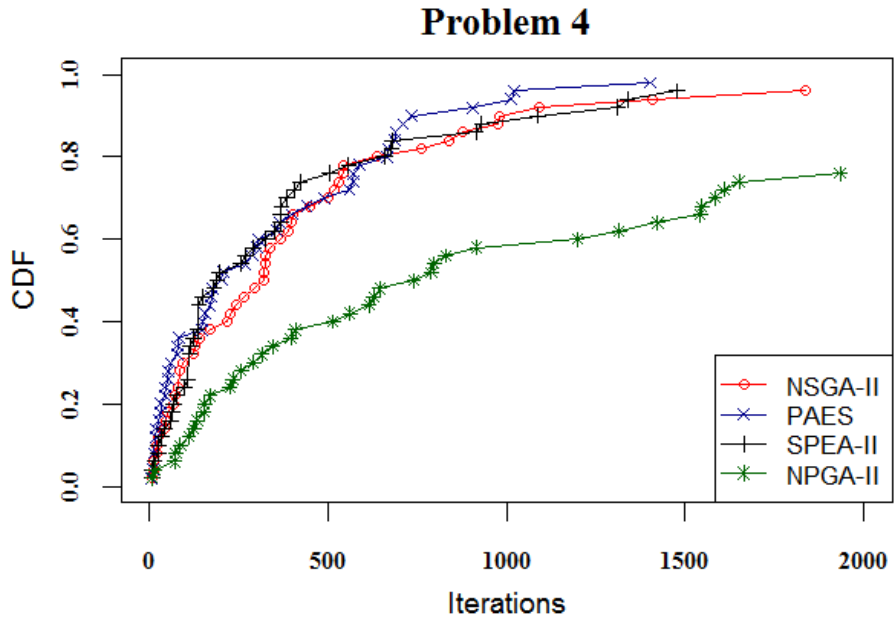

**Figure 4:** CDF of 4 MOEAs for Problem 4

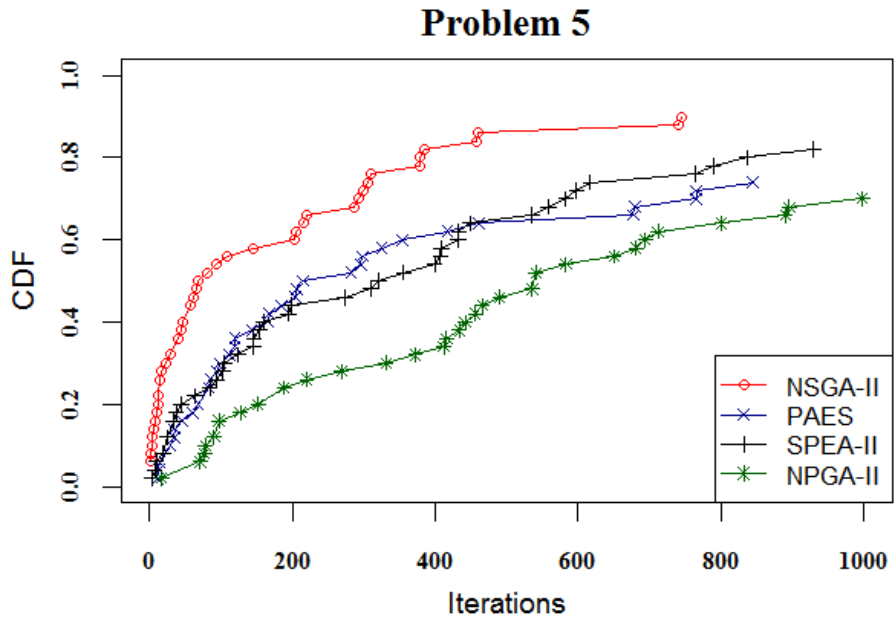

**Figure 5:** CDF of 4 MOEAs for Problem 5

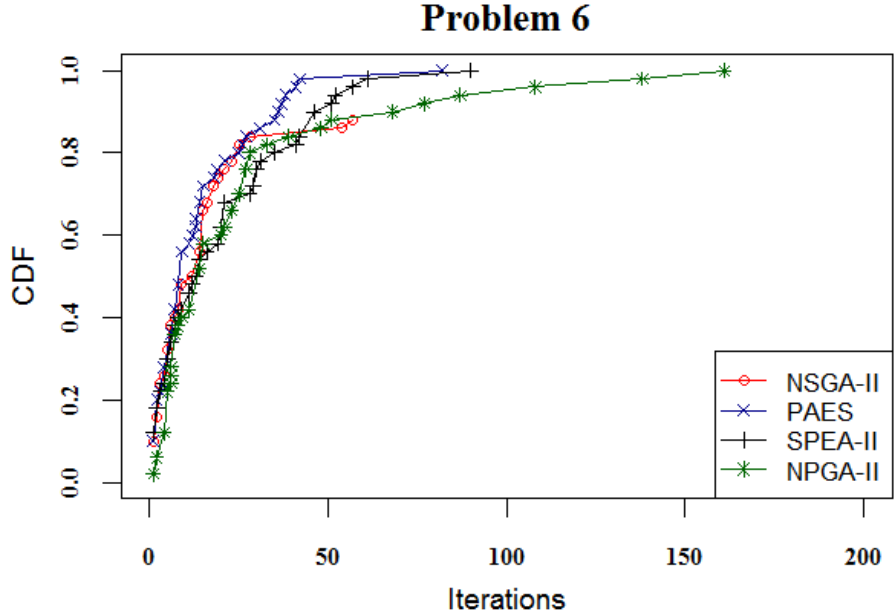

**Figure 6:** CDF of 4 MOEAs for Problem 6

**Table 2**: Mean and Variance of the iterations for the six problems

| Problem | NSGA-II | | SPEA-II | | PAES | | NPGA-II | |
|---|---|---|---|---|---|---|---|---|
| | Mean | Variance | Mean | Variance | Mean | Variance | Mean | Variance |
| Problem 1 | 1800 | 1679159 | 1583 | 1481943 | 1575 | 1452351 | 1999 | 1747775 |
| Problem 2 | 1091 | 1028848 | 1252 | 1365382 | 1479 | 1179406 | 1392 | 1126270 |
| Problem 3 | 1123 | 572831 | 992 | 581708 | 808 | 577827 | 1012 | 809507 |
| Problem 4 | 642 | 321288 | 385 | 154736 | 340 | 109901 | 361 | 147748 |
| Problem 5 | 155 | 34635 | 295 | 86506 | 231 | 51919 | 402 | 68844 |
| Problem 6 | 13 | 147 | 21 | 400 | 15 | 237 | 26 | 1163 |

Statistical analysis in Table 2 shows that for problem 1, 3 and 4 PAES takes less number of iterations to reach the termination criteria compared to other three MOEAs. For problem 2 and 5, NSGA-II takes less number of iterations compared to PAES, NPGA-II, and SPEA-II. Also, it shows expected variations in the outcome with the changing problem size. Moreover, it is important to compare the number of success amongst the 50 trials for each MOEA to reach the termination criteria. Figures 1-6 reveals that the performance of the all four MOEAs remained competitive for the six benchmark instances considered. More specifically, for problem 1, 3 and

4 PAES perform well compared to other three MOEA. For problem 2 and 5, NSGA-II performs well compared to other three MOEAs.

## 5. Algorithm Portfolio

In this section, we present the design of algorithm portfolio and its implementation on DTCTP.

### 5.1. Algorithm Portfolio Design

The deviation in the performance of the four MOEAs for DTCTP is the key to portfolio realization. Making use of the obtained variations, different portfolios have been conceived, and analysis has been carried out. These experiments were performed with portfolios containing 2 and 4 algorithms on 2 and 4 processor systems. The detailed list containing the information about the algorithms embedded in various portfolios is provided in Table 3.

The following criteria have been considered while designing these portfolios:

- Those algorithms were combined and placed in same portfolios, which showed varied performance.
- Strategies utilizing the external memory for storing non-dominated particles are tied with the strategies that store such particles in the current population.
- Portfolios were designed keeping in mind the requirement to analyze the relative performance of various strategies utilized for maintaining diversity in the population.

**Table 3**: Portfolio Design

| No. of Algorithms | Case | 2 Processors | | 4 Processors | | | |
|---|---|---|---|---|---|---|---|
| 2 | 1 | NSGA-II | PAES | NSGA-II | PAES | - | - |
|  | 2 | NPGA-II | SPEA-II | NPGA-II | SPEA-II | - | - |
|  | 3 | SPEA-II | NSGA-II | SPEA-II | NSGA-II | - | - |
|  | 4 | SPEA-II | PAES | SPEA-II | PAES | - | - |
| 4 | 1 | - | - | NSGA-II | PAES | NPGA-II | SPEA-II |

Similarly, the relative advantage that could be obtained with various Pareto diversity, maintaining strategies is also considered. Aforementioned criteria are clearly evident in the portfolio design utilized. The inferior performance of NPGA-II compared to other meta-heuristics lowers the expectancy of competitive performance of the portfolios containing it. In addition, the similarity of its operators with NSGA-II, and much better performance by NSGA-II limits its utilization (as is evident from the lowest number of its portfolios) in further portfolio analysis.

### 5.2. Implementation of Portfolio Approach

In practice, one may implement the portfolio following a few steps. First, a set of portfolio cases for different processors and its constituent algorithm should be identified. In this paper, we developed three portfolio cases, namely, 2 Algorithm-2 Processor (2A-2P), 2 Algorithm-4 Processor (2A-4P) and 4 Algorithm-4 Processor (4A-4P) (See Table 3). We then identified the constituent algorithm for each of these portfolio cases, as given in Table 3. When choosing the constituent algorithm, an intuition is that they should be more or less complementary. The constituent algorithms are combined and placed in the portfolio based on their performances on all benchmark problems. As will be shown in our numerical results interpretation, choosing complementary constituent algorithms leads to better performances with the same algorithms exploit on the different parallel processors on the same benchmark problems. More specifically, the constituent algorithms should not only employ different operators, but also exhibit different behaviors on problem sets. For second steps, algorithm portfolio framework should be highly capable of accommodating existing algorithms. However, due to the fact that some existing algorithms might have their own configuration, they merit little bit attention when incorporated into the portfolio framework. In the present portfolio, some algorithms such as NSGA-II are highly capable of providing better solution quality beside its more run time complexity. This helps respective portfolio terminates quickly if the status of the solution has reached predefined termination criteria.

In this paper, *the termination criteria for the each portfolio run was set to be the performance level within 10% of the best-obtained AQ (average quality) metric value*. Average quality (AQ) is used as standard criteria here to measure the quality of the obtained Pareto front of algorithm portfolio. More details of this issue are given in section 3. In the final step of algorithm portfolio, a number of iterations will be recorded after the termination criterion is satisfied. A number of iterations are taken as criteria to evaluate the computational efficiency of the each portfolio case.

Before running a portfolio, part of our work is to allocate population size, generation size and other parameter values for each constituent algorithms of the portfolio. Their values should be always same to evaluate the performances of each portfolio case.

## 6. Results and Discussion

In order to perform the analysis, six benchmark instances are considered with varying problem size and complexity. Simulations were performed with various combinations of algorithms and processor systems. The algorithm combinations embedded in different portfolio cases is given in Table 3. Various combinations of algorithms are implemented in parallel on 2 and 4 processor systems which are detailed in coming subsection. These combinations or portfolios are represented by a '/' notation. For example, the symbol 1/1 represents 2 Algorithms-2 Processor portfolio where the first algorithm is run on first processor whiles the other one on the second processor. Similarly, 4/0/0/0 represents 4 Algorithms-4 Processor portfolios in which four copies of the first algorithm (Table 3) were run on all the four processors. Each combination was evaluated on the basis of its performance averaged over 50 independent runs.

### 6.1. Experimental Runs

This subsection contains the experimental results obtained by considering various combinations utilized. An average number of iterations required by the portfolio to reach the desired quality level (10 % of the best) of the Pareto-optimal front is again taken as the performance criteria. Different combinations of algorithms to be run on the processors and utilized in the experiment are given as follows**:**

- **2 Processor System**: Only three possible combination of algorithms exists in this case, i.e. [2/0], [1/1] and [0/2] which are thoroughly analyzed. All the four cases were simulated with the aforementioned algorithm combinations and the obtained results are detailed in Figure 7.
- **4 Processor System** The analysis with 4-Processors system is carried out in two parts viz. 2 algorithms cases and 4 algorithms cases. In the case of 2 algorithms, the possible cases are [4/0], [3/1], [2/2], [1/3] and [0/4]. However, in 4 algorithms case, the numbers of possible combinations are extremely large. In order to reduce the computational burden due to a large number of possible portfolio combinations, authors have restricted the number of combinations utilized to 16. These selected portfolio cases are provided in Table 4 and the results obtained with two and four processors systems are shown in Figures 8 and 9 respectively.

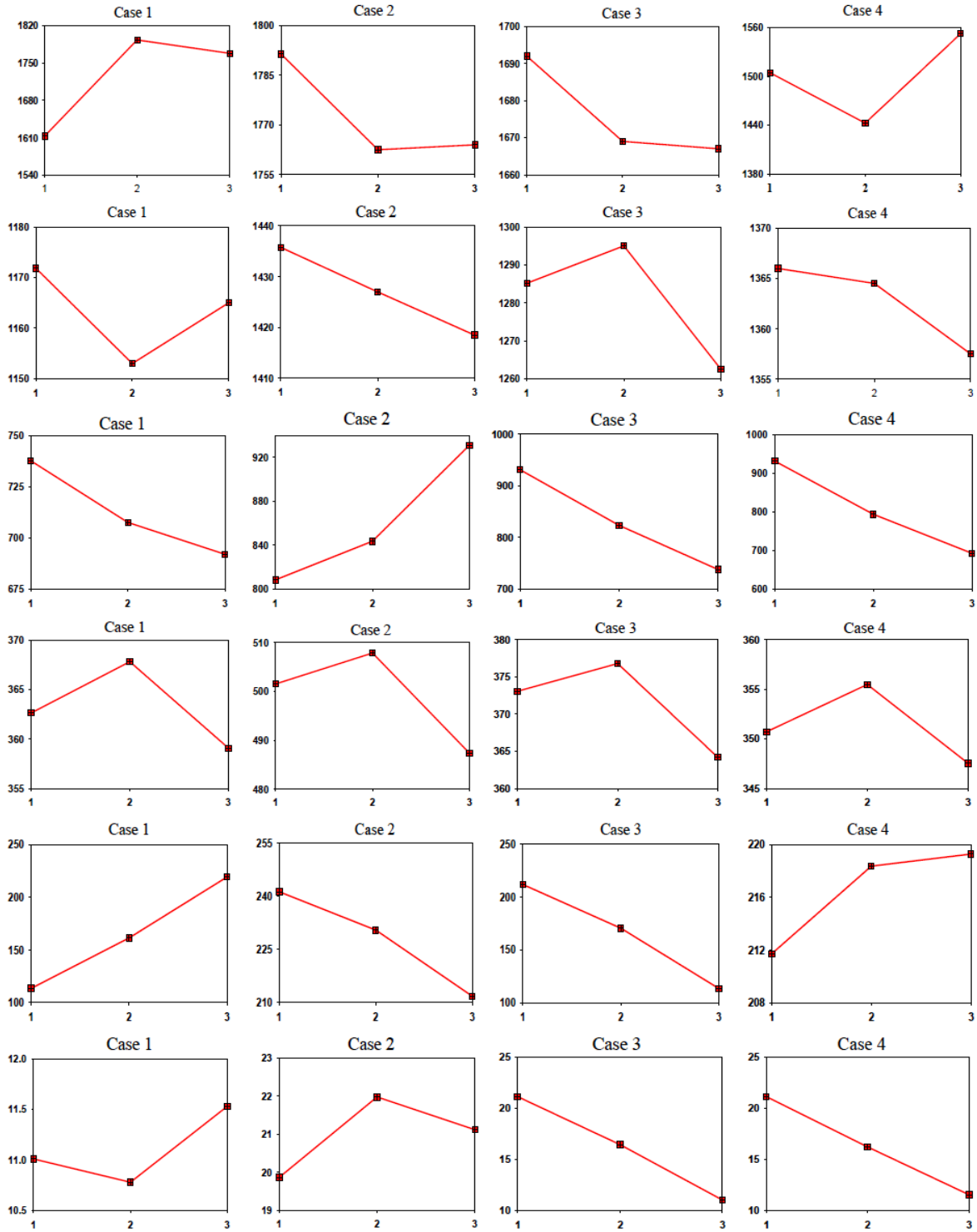

**Figure 7:** Results for different portfolio cases of 2 Algorithms-2 Processors (2A-2P) system. The six rows present results for the six benchmark problems and the columns are corresponding to different algorithms selected. Y-axis is scaled to a number of iterations; X axis denotes the portfolio cases explored.

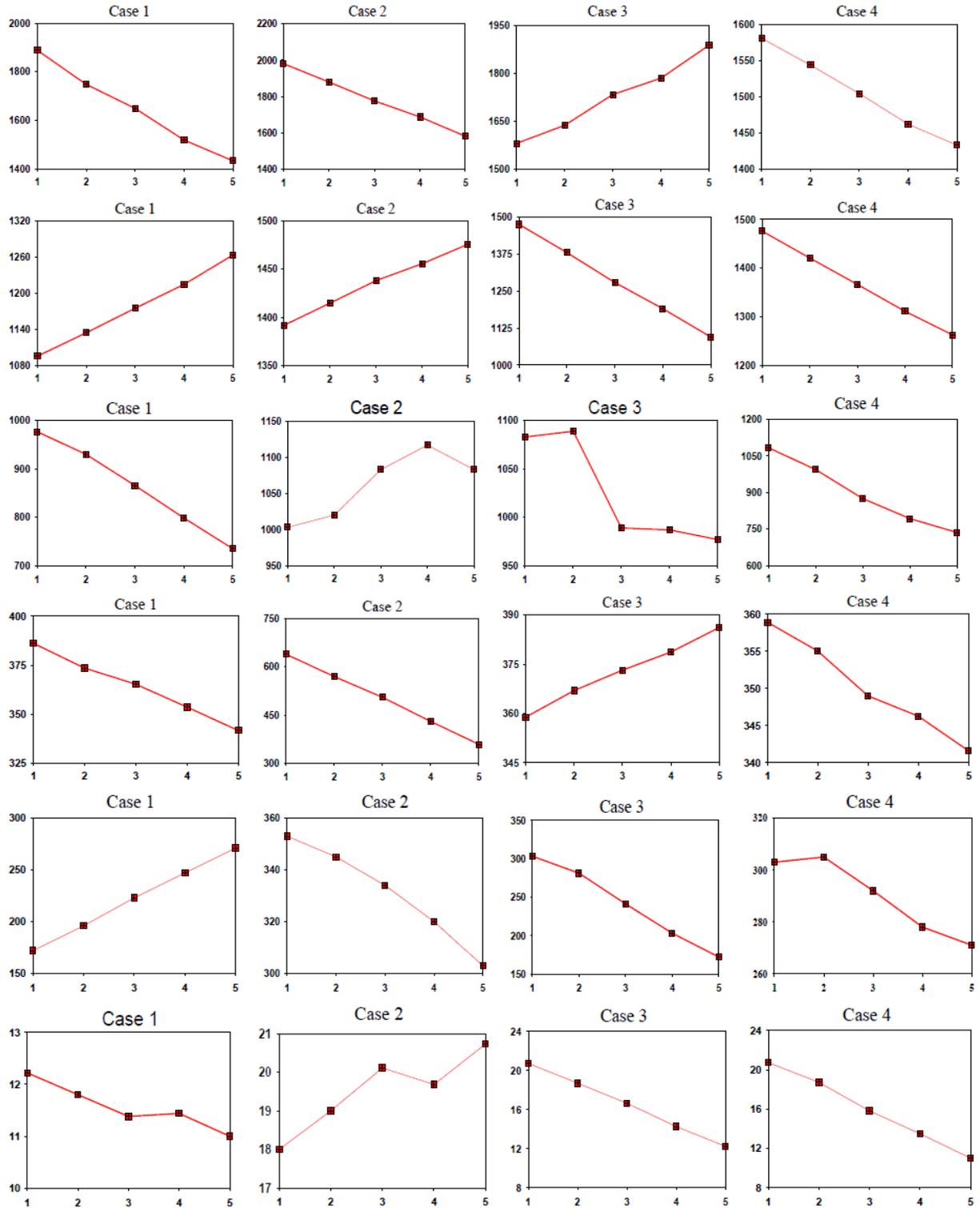

**Figure 8:** Result for different portfolio cases of 2 Algorithms- 4 Processors (2A-4P) system. The six rows present results for six benchmark problems and the column are corresponding to a different algorithm selected. Y-axis is scaled to a number of iterations; X axis denotes the portfolio cases explored.

**Table 4:** Cases investigated with 4 Algorithms-4 Processors System

| Case | 4A- 4P | Case | 4A- 4P |
|------|--------|------|--------|
| 1 | 4/0/0/0 | 9 | 0/3/1/0 |
| 2 | 3/1/0/0 | 10 | 0/4/0/0 |
| 3 | 3/0/1/0 | 11 | 1/0/3/0 |
| 4 | 3/0/0/1 | 12 | 1/2/0/1 |
| 5 | 2/1/1/0 | 13 | 0/0/4/0 |
| 6 | 2/1/0/1 | 14 | 1/1/0/2 |
| 7 | 1/0/1/2 | 15 | 0/1/0/3 |
| 8 | 0/2/1/1 | 16 | 0/0/0/4 |

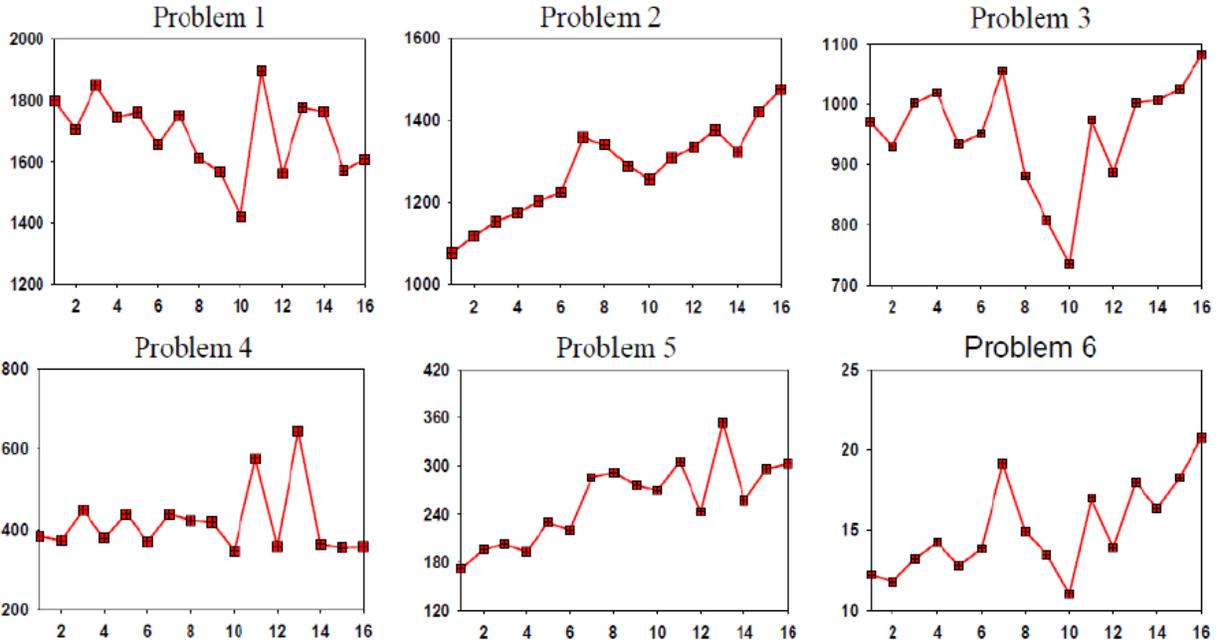

**Figure 9:** Results for different Portfolio cases of 4 Algorithms-4 Processors (4A-4P) system

## 6.2. Interpreting the Results

Having had the experimental results, the task to be accomplished is concerned to portfolio assessment. Generally, processor availability for the parallel runs is limited in an organization; hence the selection strategy has to be executed separately for each of the two types of systems (2 Processors, 4 Processors) under consideration. The varying performance of different algorithms, and eventually the portfolios, poses enough challenge to select the best strategy among the instances explored in order to get the quality solution with minimum risk. Analytical Hierarchy Process (AHP) has long been utilized as a tool to decision makers for selecting the best alternative from the given alternatives in such situations. AHP employs hierarchical pairwise comparison to induce the weights of alternatives through their pairwise comparison.

This paper also addresses the selection of best portfolio problem from an AHP perspective. Each type of portfolio is recognized as an alternative for the particular processor system to which the portfolio belongs to and the results over six different problems are considered as of different attributes. First, a matrix A is constructed which is defined as follows,

$$A = \begin{pmatrix} a_{11} & a_{12} & \cdots \\ a_{21} & a_{22} & \cdots \\ \vdots & \vdots & \ddots \end{pmatrix} \tag{17}$$

Where, the column represent attribute $\vartheta$, $\vartheta = 6$ in the underlying case (corresponding to problem 1-6); row represent the alternatives (portfolio) explored; and $a_{ij}$ represent the normalized the objective value for the experiment characterizing by set $(i, j)$, i.e.

$$a_{ij} = \frac{OBJ_{ij}}{\sum_j \{OBJ_{ij}\}} \tag{18}$$

where, $OBJ_{ij}$ represent the objective value based on which the solution quality is to be measured. In this case, $OBJ_{ij}$ symbolizes the average number of generation required by the particular test alternative.

Thereafter a priority matrix $\Omega^i_{\vartheta \times \vartheta}$ is calculated for each attribute $i$, where

$$\Omega^i_{jk} = \frac{a_{ij}}{a_{ik}} \tag{19}$$

The associated weight vector $W^i$ is then calculated for each attribute $i$ by taking geometric mean for the row corresponding to matrix $\Omega^i$

$$W^i_{1 \times \vartheta} = \{w^i_j\}$$

$$= \sqrt[\vartheta]{\prod_k^{\vartheta} \Omega^i_{jk}} \tag{20}$$

The calculation of weight vector is followed by their normalization, thus, $\vartheta$ normalized priority vector $PV_s$ is obtained as

$$PV_j^i = \frac{w_j^i}{\sum_{j=1}^{\vartheta} w_j^i} \tag{21}$$

The weighted sum of six priority vectors is now obtained that gives the relative rank vector $RV$ for the alternatives. Here, priority weights are taken as 1/6 to ensure equal weight to all attributes. For the two processor system, 8 portfolio combinations (due to a repeat of several portfolios) have been evaluated. Similarly, for 4-Processors systems, 16 combinations corresponding to both 2 algorithms and 4 algorithms cases are evaluated.

The final rank vectors and ranks for different alternatives are provided in Tables 5, 6 and 7. As is evident, the aforementioned strategy provides a relative ranking of the various portfolio alternatives available.

**Table 5:** Priority vectors and corresponding rank for the portfolio cases of 2 Algorithms and 2 Processors system

| 2 Algorithms - 2 Processors (2A-2P) System | | | |
|---|---|---|---|
| Algorithm selected | Portfolio | Priority | Rank |
| NSGA-II – PAES | **2/0** | **0.1666** | **3** |
|  | **1/1** | **0.1623** | **1** |
|  | 0/2 | 0.1723 | 5 |
| NPGA-II – SPEA-II | 2/0 | 0.1785 | 7 |
|  | 1/1 | 0.1974 | 12 |
|  | 0/2 | 0.1886 | 9 |
| SPEA-II – NSGA-II | 2/0 | 0.1913 | 10 |
|  | **1/1** | **0.1634** | **2** |
|  | 0/2 | 0.1766 | 6 |
| SPEA-II – PAES | 2/0 | 0.1820 | 8 |
|  | 1/1 | 0.1924 | 11 |
|  | 0/2 | 0.1683 | 4 |

For the 2 Algorithms-2 Processors (2A-2P) System, the portfolios comprising of algorithms NSGA-II and PAES characterized by (1/1) system is found to be the best. One of the possible reasons for this might be the complementary performance of both the algorithms on the problem scenario considered. As is evident from the individual performance profiles (Figures 1-

6), in general, the performance of PAES was good for all instances. On the contrary, the performance of NSGA-II was good for large and medium sized instances while considerably poor for the small sized problem. The results obtained with portfolios embedding NSGA-II are generally better thereby supporting our early assumptions for it's increased the number in the considered portfolios. It is also evident that NPGA-II, which performed poorly when considered individually, continued the show with the portfolios embedding it resulting in worst case performance.

**Table 6:** Priority vectors and corresponding rank for the portfolio cases of 2 Algorithms - 4 Processors system

| 2 Algorithms - 4 Processors (2A-4P) System | | | |
|---|---|---|---|
| Algorithm | Portfolio | Priority | Rank |
| NSGA-II – PAES | 4/0 | 0.1712 | 4 |
| | **3/1** | **0.1667** | **3** |
| | **2/2** | **0.1631** | **1** |
| | 1/3 | 0.1784 | 7 |
| | 0/4 | 0.2464 | 6 |
| NPGA-II – SPEA-II | 4/0 | 0.2185 | 14 |
| | 3/1 | 0.2213 | 15 |
| | 2/2 | 0.2313 | 20 |
| | 1/3 | 0.2286 | 19 |
| | 0/4 | 0.2256 | 17 |
| SPEA-II – NSGA-II | 4/0 | 0.2234 | 16 |
| | 3/1 | 0.2133 | 13 |
| | 2/2 | 0.1934 | 10 |
| | 1/3 | 0.1833 | 8 |
| | **0/4** | **0.1634** | **2** |
| SPEA-II – PAES | 4/0 | 0.2266 | 18 |
| | 3/1 | 0.2076 | 12 |
| | 2/2 | 0.1986 | 11 |
| | 1/3 | 0.1867 | 9 |
| | 0/4 | 0.1767 | 5 |

For the 2 Algorithms - 4 Processors (2A-4P) System, the portfolio comprising of algorithms NSGA-II and PAES characterized by (2/2) system is found to be the best. From the result, it is clear that running the algorithm on the parallel processor is also found to be the best strategy when harmonizing portfolio. Similarly, for the 4 Algorithms case, the portfolio comprising of algorithms NSGA-II and PAES characterized by (3/1/0/0) system is found to be the best-suited strategy.

In fact, the finding is attributed to the element of randomness inherited in the search procedure. Randomness here does not mean uncertainty rather it refers to the ability of the heuristic to vary its performance based on the initial random seed. However, the reason can be deduced on the basis of performance profiles obtained with the single processor system (Figures 1-6).

**Table 7:** Priority vectors and rank for the portfolio cases of 4 Algorithms - 4 Processors (4A-4P) system

| 4 Algorithms - 4 Processors (4A-4P) System | | | |
|---|---|---|---|
| Algorithm | Portfolio | Priority | Rank |
| NSGA-II – PAES - NPGA-II - SPEA-II | **4/0/0/0** | **0.1587** | **2** |
| | **3/1/0/0** | **0.1567** | **1** |
| | 3/0/1/0 | 0.1631 | 4 |
| | 3/0/0/1 | 0.1667 | 6 |
| | 2/1/1/0 | 0.1646 | 5 |
| | 2/1/0/1 | 0.1723 | 7 |
| | 1/0/1/2 | 0.1924 | 15 |
| | 0/2/1/1 | 0.1789 | 10 |
| | 0/3/1/0 | 0.1766 | 9 |
| | **0/4/0/0** | **0.1616** | **3** |
| | 1/0/3/0 | 0.1833 | 13 |
| | 1/2/0/1 | 0.1746 | 8 |
| | 0/0/4/0 | 0.1889 | 14 |
| | 1/1/0/2 | 0.1813 | 11 |
| | 0/1/0/3 | 0.1867 | 13 |
| | 0/0/0/4 | 0.1937 | 16 |

As is evident, NSGA-II shows higher CDF in regions of less number of iterations thereby empirically suggestive of the aforementioned fact. Similarly, other interesting and supporting conclusions (for portfolio implementation) can be drawn from the analysis. However, before the implementation of this concept to the practical scenario, a much detailed experimentation is required. This comprises of a much exhaustive list of algorithms and an augmented use of the processors.

## 7. Conclusion and Future Research

This paper proposes a portfolio approach to algorithm selection for discrete time-cost trade-off problem in the multi-objective environment. The proposed algorithm portfolio containing NSGA-II, PAES, NPGA-II and SPEA-II is used to minimize the associated risk related to the selection of algorithms. Six benchmark instances of DTCTP of varying dimension and complexities are included to analyze the performances of a portfolio approach. The four algorithms are parallelly processed on two and four processors system. The results showed an insight, suggesting the application of portfolios to select the best algorithm for computationally expensive multi-objective optimization problems. Furthermore, the result shows a considerable decrease in the computational time by the parallel processing of algorithms.

As a perspective for future work, the much detailed analysis is needed prior to any suggestion for algorithm portfolio usage. However, our preliminary results are encouraging new directions for implementation of the algorithm portfolio on different multi-objective decision-making problems.